\documentclass[journal]{IEEEtran}
\usepackage{cite}
\usepackage{amsmath,amssymb,amsfonts}
\usepackage{comment}
\usepackage{graphicx}
\usepackage{textcomp}
\usepackage{booktabs}
\usepackage{array}
\usepackage{enumitem}
\usepackage{url}
\usepackage{xcolor}
\usepackage[hidelinks]{hyperref}

\newcommand{\orcid}[1]{\href{https://orcid.org/#1}{\raisebox{-0.15ex}{\includegraphics[height=1.6ex]{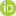}}}}

\makeatletter
\newsavebox{\@figshimbox}
\def\Figure{\@ifnextchar[{\@Figure}{\@Figure[!t]}}
\def\@Figure[#1]{\@ifnextchar[{\@@Figure[#1]}{\@@Figure[#1][scale=1]}}
\def\@@Figure[#1][#2]#3#4{%
  \sbox\@figshimbox{\includegraphics[#2]{#3}}%
  \ifdim\wd\@figshimbox<\columnwidth
    \begin{figure}[#1]\centering\usebox\@figshimbox\caption{#4}\end{figure}%
  \else
    \begin{figure*}[#1]\centering\usebox\@figshimbox\caption{#4}\end{figure*}%
  \fi}
\makeatother

\let\PARstart\IEEEPARstart

\makeatletter
\def\@IEEEsectpunct{\ \,}
\makeatother

\graphicspath{{../}}

\begin{document}

\title{A Tutorial on Prompt Engineering:\\From Messy Thoughts to AI Workflows}

\author{Erfan~Loweimi\,\orcid{0000-0002-8761-021X}$^{1,2}$,
Hadi~Daneshvar\,\orcid{0000-0002-2727-7646},
Samira~Loveymi\,\orcid{0000-0002-7431-7671}$^{3}$,
Samir~Ouelha\,\orcid{0000-0002-8251-2282}$^{1}$,
Zhengjun~Yue\,\orcid{0000-0002-1101-549X}$^{4}$,
Hajar~Mozaffar\,\orcid{0000-0002-6400-5260}$^{5}$,
Saturnino~Luz\,\orcid{0000-0001-8430-7875}$^{2}$
\thanks{$^{1}$Cisco, UK}
\thanks{$^{2}$Centre for Medical Informatics, Usher Institute, University of Edinburgh, UK}
\thanks{$^{3}$Department of Computer Engineering, Ahvaz Campus, Islamic Azad University, Iran}
\thanks{$^{4}$SLAI \& CUHK-SZ, China}
\thanks{$^{5}$Business School, University of Edinburgh, UK}
\thanks{Corresponding author: Erfan Loweimi (e-mail: eloweimi@cisco.com).}}

\markboth{Loweimi \MakeLowercase{\textit{et al.}}: A Tutorial on Prompt Engineering}
{Loweimi \MakeLowercase{\textit{et al.}}: A Tutorial on Prompt Engineering}

\maketitle

\begin{abstract}
This paper treats prompt engineering as a discipline for turning informal human intent into structured AI work specifications. It develops the practice as a sequence of reusable design moves: define the work, construct only the context the answer depends on, choose a role, or a moderated panel of roles, as an attention lens, and state affirmative quality targets, reserving prohibitions for hard boundaries. To keep prompts lean, it adapts two classical principles, Occam's razor and Chekhov's gun, so that every instruction earns its place. For consequential tasks, it adds structured critique through steelmanning and premortems, followed by verification and, where tools or multi-step actions are involved, agentic operating loops with explicit boundaries and escalation. Aimed at a general readership, this tutorial is not a benchmarking study; it offers a practical, technically grounded path from casual prompting to disciplined AI workflow design, illustrated by a worked example that carries one task from a weak prompt to a strong specification. The framework is presented as principles and checklists that remain useful as models and tools change. The strongest prompt is rarely the longest prompt; it is the one that makes desired behaviour, required sources and checks, and success criteria unmistakable.
\end{abstract}

\begin{IEEEkeywords}
Agentic AI, context engineering, generative artificial intelligence, human-AI interaction, large language models, problem formulation, prompt engineering
\end{IEEEkeywords}

\IEEEpeerreviewmaketitle

\section{Introduction}
\label{sec:introduction}
\PARstart{M}{odern} large language model (LLM) interfaces make it easy to get a workable answer from a casual request. A technical treatment of prompt engineering should begin from that fact. Weak prompts can work; they often do. The better question is what changes when the task becomes ambiguous, repeated, high-consequence, multi-step, or agentic. In those settings, the prompt stops being merely a request and becomes a specification for work: a translation of informal human intent into terms an AI system can act on. In the agentic AI era, prompt engineering also requires designing context, constraints, tools, critique loops, verification steps, and human oversight so that AI systems can perform useful work within explicit, reviewable boundaries.

This paper therefore treats prompt engineering as a problem-formulation and workflow-design discipline. A good prompt makes the task, constraints, evidence requirements (what sources or checks the answer should rely on), and success criteria clearer to the model and to the human reviewer. Clarity is not correctness, however: a clear prompt can still produce a wrong answer. Prompt engineering must be paired with verification, source checking, and human judgment.

A strong foundation model can often recover from a vague request. Prompt engineering matters because it helps the user get the intended result with less waste, fewer hidden assumptions, and clearer review points. In this sense, prompt engineering can be read as an optimisation problem, where the quantities to be reduced are turns, tokens, retries, repairs, side problems, and downstream errors. The paper does not measure those quantities; it argues for the design choices that reduce them by making assumptions, criteria, and review points explicit.

\paragraph{Scope and contributions.}
The paper makes six practical contributions:
\begin{enumerate}[leftmargin=*]
    \item a work-specification view of prompts rather than a search for clever phrasing;
    \item a structure for converting vague requests into goals, context, constraints, output requirements, and review criteria;
    \item a treatment of role playing as an attention lens instead of a vague persona, extended to moderated panels;
    \item a distinction between affirmative quality targets and negative guardrails;
    \item critique through \textit{steelmanning} and \textit{premortem} reasoning, with verification, as mechanisms for tasks where error cost justifies added structure;
    \item an agent-prompt anatomy that defines tools, action boundaries, observations, stop rules, escalation, and reporting.
\end{enumerate}

Taken together, these contributions treat prompt engineering as disciplined problem formulation: informal intent is converted into an explicit work specification, which a model or agent then executes, and the result feeds a review loop that refines the specification. What must be made explicit is the goal, context, criteria, constraints, and verification, stated clearly enough that both the model and the human can examine them. Figure~\ref{fig:prompt-specification} summarises this view.

\Figure[!t][width=\textwidth]{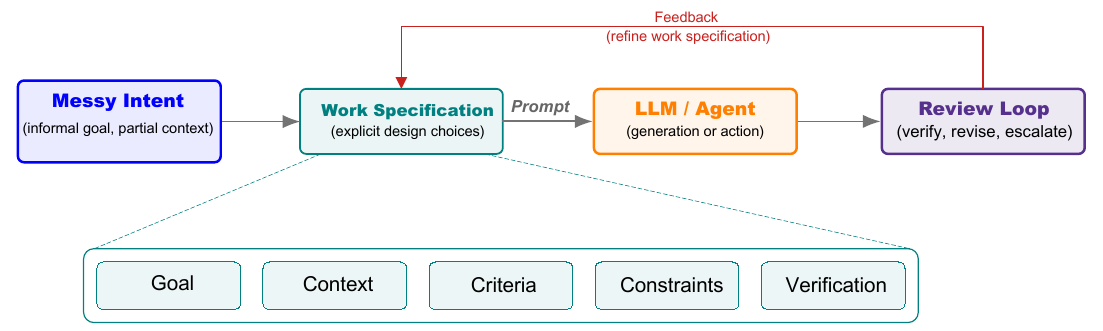}{Prompt engineering as work specification: informal intent becomes explicit design choices, executed by an LLM or agent and improved through review.\label{fig:prompt-specification}}

The rest of the paper is organised as follows. Section~\ref{sec:related} reviews related work. Section~\ref{sec:methodology} explains how the framework was constructed. Section~\ref{sec:implementation} develops the framework step by step. Section~\ref{sec:synthesis} summarises the framework as checklists and design principles. Section~\ref{sec:discussion} discusses boundaries, risks, and practical implications. Section~\ref{sec:conclusion} concludes the paper.

\section{Related Work}
\label{sec:related}

Prompt engineering builds on several lines of work. \textit{Few-shot} prompting showed that large language models can adapt to tasks from examples in the prompt \cite{brown2020fewshot}. \textit{Prompt-based learning} surveys connect this practice to a broader family of methods that formulate natural language processing tasks as text prompts \cite{liu2023pretrainpromptpredict}. Instruction-following work, including InstructGPT, shifted attention from task completion alone to models that better follow user intent \cite{ouyang2022instructgpt}. Prompt-pattern catalogues describe reusable interaction patterns for LLMs \cite{white2023promptpatterns}, while systematic surveys organise prompting techniques across reasoning, generation, tool use, and evaluation \cite{schulhoff2024promptreport,sahoo2024survey}. Automated prompt optimisation methods, including Automatic Prompt Engineer \cite{zhou2023ape} and automatic long-prompt engineering \cite{hsieh2023longprompts}, show that prompt design can itself be treated as an optimisation problem.

Practitioner guidance adds a complementary layer. Google's whitepaper presents practical prompt components, examples, and prompting techniques for Gemini-style systems \cite{boonstra2024prompt}. Google Cloud's guide frames prompt engineering as the practice of crafting inputs that guide generative AI systems towards useful outputs \cite{googlecloud2026prompt}. OpenAI's prompting guidance emphasises clear instructions, context, examples, and iterative refinement \cite{openai2026prompting,openai2026promptengineering}. Anthropic's guidance discusses how prompting can influence extended reasoning behaviour and verification in more demanding tasks \cite{anthropic2026extendedthinking}. Weng's overview connects prompting to in-context learning, reasoning prompts, and tool use \cite{weng2023prompt}.

Prompting also intersects with work on reasoning and action. \textit{Chain-of-thought} prompting structures intermediate reasoning \cite{wei2022chain}; \textit{self-consistency} samples multiple reasoning paths before selecting an answer \cite{wang2023selfconsistency}; \textit{Tree of Thoughts} frames problem solving as search over intermediate states \cite{yao2023tree}; and \textit{ReAct} combines reasoning traces with tool actions and observations \cite{yao2023react}. These works matter for agentic AI because the prompt no longer asks only for an answer; it may define a loop of reasoning, acting, observing, and updating. Engineering guidance on agents similarly recommends clear workflows and control flow before adding autonomy \cite{anthropic2024agents}.

\textit{Retrieval-augmented generation} (RAG) grounds model outputs in supplied or retrieved material rather than only in what the model learned during training \cite{lewis2020rag}. Context-engineering work generalises this idea by treating instructions, retrieved documents, tools, memory, and state as parts of the information payload available to the model \cite{mei2025contextengineering}. Security work shows why this payload must be handled carefully: \textit{prompt injection} arises when untrusted text is mixed with instructions in ways that can manipulate the system \cite{willison2022promptinjection}. \textit{Indirect prompt injection} extends the problem to text an application retrieves rather than text a user types, and has been demonstrated against deployed systems including search assistants and code-completion engines \cite{greshake2023injection}. OpenAI guidance therefore emphasises constraining what untrusted content can affect \cite{openai2026promptinjection}, a principle consistent with broader AI risk-management and LLM application security guidance \cite{nist2023airmf,owasp2025llm}.

These works differ from the present paper in what they are organised around. Surveys and catalogues are organised by technique: they list prompting methods and describe when each has been reported to help \cite{schulhoff2024promptreport,sahoo2024survey,white2023promptpatterns}. Vendor guides are organised by product: each explains how to prompt one company's system. Context engineering is organised by the information given to the model \cite{mei2025contextengineering}, and comes closest to the present paper; the difference is that this information is only part of what a work specification must settle, alongside success criteria, critique, verification, and human oversight. This paper is organised around the work being delegated. That starting point raises three questions a list of techniques does not answer: what a prompt must make explicit before any technique is chosen, how much structure a task justifies, and what the prompt must specify once the model can act rather than only answer.
Answering them is the paper's contribution: the specification itself, and the rule for deciding how much of it a task needs.

\section{Methodology}
\label{sec:methodology}

This paper is a tutorial built by synthesising existing work. The methodology has two purposes: first, to retain the technically grounded ideas in prompting research and practitioner guidance; second, to arrange those ideas into a usable design sequence for prompt construction. The synthesis drew on two kinds of material:

\begin{itemize}[leftmargin=*]
    \item research on prompt-based learning, instruction following, chain-of-thought prompting, self-consistency, Tree of Thoughts, ReAct, retrieval-augmented generation, automatic prompt optimisation, and context engineering;
    \item practitioner and standards guidance from Google, OpenAI, Anthropic, the National Institute of Standards and Technology (NIST), and the Open Worldwide Application Security Project (OWASP), together with wider AI engineering practice, covering prompting, agents, tool use, evaluation, and security.
\end{itemize}

The resulting framework also uses general problem-solving principles, such as role-based review, steelmanning \cite{aikin2011strawmen}, premortems \cite{klein2007premortem}, \textit{Occam's razor} \cite{wikipedia_occam,domingos1999occam}, and \textit{Chekhov's gun} \cite{wikipedia_chekhov,britannica_chekhov}, as design lenses. They guide how a prompt is written; they are not evidence that it works.

\paragraph{Scope.}
The material comprises peer-reviewed work from natural language processing and machine learning venues, preprints already in common practical use, and vendor and standards documentation, cited with its access date. A source was included when it either introduced a prompting or agent technique that remains in use across model generations, or gave design guidance specific enough to be turned into a decision rule. Two kinds of material were excluded: results whose validity is tied to one model release or one provider, for the reason argued in Section~\ref{sec:discussion}, and guidance that repeats advice already covered by an included source.

The process followed four steps.

\paragraph{Step 1: Extract the core claims.}
The main claims were identified from recurring themes across the reviewed material: prompts are work specifications; prompt engineering is an optimisation discipline \cite{zhou2023ape,hsieh2023longprompts}; agentic AI turns prompts into workflow definitions \cite{yao2023react,anthropic2024agents}; and good prompting depends on principles over magic wording \cite{sprague2025cot,sclar2023sensitivity}.

\paragraph{Step 2: Group the ideas into design activities.}
Those ideas were reorganised into six activities: task definition, context construction, instruction design, reasoning and critique, agent workflow specification, and evaluation. This grouping creates a progression from basic prompting to agentic workflow design. It also gives the framework a repeatable working sequence: first define the work, then build the context, then write the instructions, then guide the reasoning, then specify the agent workflow, and finally evaluate the result.

\paragraph{Step 3: Remove redundancy and unsupported claims.}
A claim was kept only when it changed what a prompt should contain, and was then restated as a decision rule. Claims that were vague, repeated another source, or held only for one model or benchmark were dropped.

\paragraph{Step 4: Select examples only when they clarify abstractions.}
The paper uses a small number of qualitative examples: a chief financial officer (CFO) role-playing example, a concise-output instruction example, an Occam/Chekhov contrast, an agent anatomy diagram, and a worked example that carries one task from a weak prompt to a full specification. These examples are included because they clarify otherwise abstract principles. The paper deliberately avoids filling the argument with prompt templates and magic wording; the aim is to support reasoning about prompts, not memorising phrases.


\section{A Framework for Prompt Engineering}
\label{sec:implementation}

\subsection{Framework Overview}

The framework turns the six activities identified in Step 2 of Section~\ref{sec:methodology} into seven design moves. Instruction design splits into three separate decisions, choosing a lens, defining the target, and cutting what does not earn its place, while evaluation shapes the next prompt rather than the current one and is treated separately in Section~\ref{sec:synthesis}. These moves recur across non-trivial prompt-engineering tasks:

\begin{enumerate}[leftmargin=*]
    \item \textbf{Name the work.} State the task, decision, or deliverable. Do not begin with style or role before the work is clear.
    \item \textbf{Build common ground.} Add only the context the answer depends on: audience, constraints, source material, domain assumptions, and current state.
    \item \textbf{Choose the lens.} Add a role only when it changes what the model should attend to. A useful role carries perspective, criteria, and responsibility.
    \item \textbf{Define the target.} Prefer affirmative instructions that specify observable behaviour: format, length, evidence requirements, examples, or acceptance criteria. Reserve prohibitions for hard limits.
    \item \textbf{Make every element earn its place.} Remove decorative complexity and keep only what shapes the output, reduces ambiguity, prevents failure, or supports verification.
    \item \textbf{Add critique and verification.} For ambiguous or high-stakes tasks, add steelmanning, stakeholder review, premortem reasoning, or explicit checks on the answer.
    \item \textbf{Specify the operating loop.} When tools or multi-step actions are involved, define allowed actions, observations, stop rules, escalation, and reporting.
\end{enumerate}

This sequence is flexible; the seven moves describe the design space: simple tasks may require only a direct prompt, while complex workflows require more explicit context, critique, and tool boundaries. Prompts worth reusing are then improved through evaluation and revision, using the prompt logs, test cases, and rubrics discussed in Section~\ref{sec:synthesis}.

\subsection{A Prompt Escalation Ladder}

A common failure mode in prompt engineering is over-engineering simple tasks. A useful prompt is appropriately specified for the task, which may or may not make it long. Added structure is a cost that has to be justified rather than a default: a meta-analysis of more than 100 studies, together with evaluations on 20 datasets and 14 models, found that chain-of-thought prompting delivers large gains mainly on mathematical and symbolic tasks, such as arithmetic and formal logic, and much smaller gains elsewhere, such as commonsense reasoning and knowledge questions \cite{sprague2025cot}. The practical question is therefore which tasks earn which kind of structure.
\begin{table*}[!t]
\centering
\caption{Prompt escalation ladder for balancing effort, rigour, and practicality.}
\label{tab:escalation}
\begin{tabular}{p{0.21\textwidth}p{0.36\textwidth}p{0.34\textwidth}}
\toprule
\textbf{Task type} & \textbf{Prompting approach} & \textbf{Reason} \\
\midrule
Low-stakes, familiar, reversible & Use a direct prompt with the desired output. & Extra structure may cost more than it improves. \\
Ambiguous or reusable & Add context, constraints, output format, and success criteria. & The prompt becomes a reusable work brief. \\
High-stakes or assumption-heavy & Add critique, steelmanning, premortem, and verification. & The goal is to expose hidden assumptions before commitment. \\
Tool-using or agentic & Add tool permissions, observation rules, stop conditions, escalation, and reporting. & The model is no longer only answering; it is operating inside a workflow. \\
\bottomrule
\end{tabular}
\end{table*}

Table~\ref{tab:escalation} gives a practical escalation ladder. Each rung keeps what the rung below asks for and adds one thing: first a direct prompt, then a reusable work brief, then critique and verification, and finally limits on what the model may do. A task moves up not merely because it is hard but because it matters: an error is costly, a flaw in a reused prompt repeats with every use, or the model can act on files, tools, or systems rather than only answer. The boundaries are judgment calls: a prompt that is used often but decides little stays low, while a single costly decision belongs higher. The ladder also runs downward; when a task turns out to be routine, drop the structure it no longer needs. The ladder balances a central tension: technical rigour asks for explicitness, while usability asks for restraint. The rule is to add structure when it reduces ambiguity, prevents failure, improves verification, or makes delegated work safer. Otherwise, keep the prompt simple.

\subsection{Prompt as Work Specification}

The first design move is to stop asking, ``What should I say to the model?'' and start asking, ``What work am I delegating?'' A prompt that only requests an answer leaves the model to infer the missing goal, audience, constraints, evidence requirements, and success criteria. In practice, the model does not stop on under-specification; it picks plausible defaults for whatever is left unsaid. Those defaults are often reasonable, but they are the model's choices, not the user's. The gap is easy to miss on a single answer and easy to feel across reused prompts, longer sessions, and agentic execution, where each silent default has more chances to disagree with intent. Knowing which of those blanks matter is itself expertise. As agents improve at delivering to a clear specification, the human's work shifts to deciding what that specification must contain, and an inexperienced user gets weaker results not because the model is weaker but because they do not know what context it needs \cite{ng2026skillsmap}. A prompt treated as a work specification makes those choices visible. Operationally, this is a re-encoding step: unstructured intent is rewritten into a compact form that both the model and the human reviewer can inspect. The central unit of prompt engineering is a brief for useful work: goal, context, input, constraints, lens, output format, verification, and escalation.

Let a prompt specification be represented as
\begin{equation}
P = \langle G, C, I, K, L, F, V, E \rangle,
\label{eq:spec}
\end{equation}
where $G$ is the goal, $C$ is context, $I$ is input material, $K$ is constraints, $L$ is the role or reasoning lens, $F$ is output format, $V$ is verification criteria, and $E$ is human oversight or escalation. This notation is not proposed as a mathematical model of LLM behaviour. It is a compact way to represent the work-brief view of prompting. Each element is a fragment of that brief. Most are written in natural language, though the output format and verification criteria may instead be given in machine-readable form when a program consumes the result.

In practice, a prompt need not instantiate every element. The appropriate level of structure depends on task ambiguity, stakes, repetition, and whether the system can take actions. The aim is not to make prompts longer; it is to remove avoidable ambiguity before the model commits to a path.

\subsection{Context Construction}

The second design move is to build common ground explicitly. Useful context names who is asking, the objective, constraints, domain, audience, stage of work, and preferred response style. Include what the model needs to work out the answer, and nothing more. Context must therefore be curated, not accumulated: added material competes for the model's attention and for space in its context window, the limited amount of text it can read at once. Models also use long inputs unevenly: performance follows a U-shaped curve, highest when the relevant material sits at the beginning (primacy bias) or the end (recency bias), and lowest when it is buried in the middle, even in models built for long contexts \cite{liu2024lostmiddle}. Padding a prompt therefore does not merely waste space; it can hide the material that mattered.

This leads to two practical rules.

\begin{enumerate}[leftmargin=*]
    \item Put what stays true across tasks in persistent project context: business plan, product definition, audience, tone, standards, and durable preferences.
    \item Put what applies only to the task at hand in the current prompt: immediate objective, fresh constraints, source material, deadline, and recent decisions.
\end{enumerate}

For agents, context construction becomes operating context. Persistent instructions may define scope, permissions, escalation rules, and what the agent must verify before acting. Temporary instructions define the immediate objective and current evidence.

\paragraph{Design check.}
Before adding context, ask whether the detail changes the answer, constrains the work, or helps verify the result. If it does none of these, remove it or keep it outside the prompt.

\subsection{Role Playing as Attention Design}

The third design move is to use role playing only when a role changes how the model approaches the task. In short, a role works as a perspective, not a persona: it tells the model what to look at and by which standards, not who to pretend to be. The examples that follow use one running task: whether a board should approve a fixed \$5M, three-year plan to expand a software-as-a-service (SaaS) business. A vague prompt such as ``Act as an expert CFO and review this plan'' may help with tone, vocabulary, and framing, but by itself it is too vague to enforce expert judgment or set clear criteria. Evidence supports this caution: a study of 162 personas across four model families and 2,410 factual questions found that adding a persona to the system prompt did not improve accuracy over a no-persona control, and that the effect of any particular persona was largely unpredictable \cite{zheng2024personas}. The role label alone does no work. The prompt must state what that label implies: whose perspective to take, which criteria to apply, and what the answer is responsible for.

\Figure[!t][width=0.95\columnwidth]{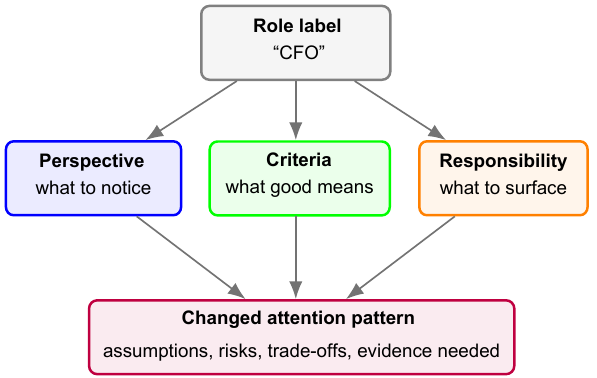}{Role playing works when it specifies perspective, criteria, and responsibility instead of merely asking for an expert persona.\label{fig:role-playing}}

The stronger CFO version is:

\begin{quote}
Review this plan from the perspective of a CFO responsible for cash discipline, downside risk, implementation cost, and measurable return on investment (ROI). Identify the assumptions that would matter most before approval.
\end{quote}

This version is stronger because it names the perspective, criteria, and responsibility, the three elements shown in Figure~\ref{fig:role-playing}. It tells the model what a CFO should attend to: cash discipline, downside risk, implementation cost, measurable ROI, and approval-relevant assumptions. The model does not become a CFO. The narrower claim is that the role changes what the model treats as important, and therefore what the answer emphasises.

\paragraph{Design check.}
When adding a role, complete the sentence: ``This role should attend to \_\_\_ (perspective), judge by \_\_\_ (criteria), and be accountable for \_\_\_ (responsibility).'' If those blanks are empty, the role is probably decorative.

\paragraph{Panels and moderation.}
A single lens extends naturally into a panel of complementary roles, but only if the panel is run to expose disagreement rather than to reach agreement. The simplest arrangement is two short rounds: one to gather views, one to test them. In the first, the roles answer independently, because roles that see each other's answers are influenced by what has already been said instead of forming an independent view, and the panel then herds towards a single position rather than examining it. In practice, issuing the first round as separate requests is easiest, since in one pass each role sees the ones before it. The second round makes that dependence deliberate: the moderator takes the sharpest disagreements and puts them back to the roles they challenge, so the panel is cross-examined rather than merely polled. Further rounds are possible, but each costs another pass and should test something the previous round left open. The moderator then reports a brief rather than a verdict: what the panel agrees on, and where it splits and why. The decision stays with the user. For the running example, the panel takes this form:

\begin{quote}
Review this expansion plan three times, independently, without adjusting any review to agree with the others: as a CFO accountable for cash discipline and measurable ROI, as a delivery lead accountable for staffing and implementation risk, and as a sales lead accountable for the demand assumptions. Then act as moderator: identify the sharpest disagreements, those that would most change the approval decision, put each back to the roles it challenges for one round of response, and report what the panel agrees on and where it splits and why. End with that brief, not a recommendation to approve or reject. (Optionally, append the three reviews so the brief can be checked against them.)
\end{quote}

\paragraph{From roles to skills.}
A recent generalisation of the role idea is the notion of \textit{skills}: packaged, reusable specifications that bundle role, scope, allowed tools, procedural steps, and example outputs into a unit the model can load on demand \cite{anthropic2025skills}. Where a role is a lens activated by prompt phrasing, a skill is an explicit package, often a small folder of instructions, templates, and code, that the model brings into context only when the task matches its purpose. Loading on demand is context construction applied to instructions: guidance the task does not need stays out of the window. Skills inherit the role-as-lens logic but formalise it as a versioned, shareable design output: instead of rewriting a strong CFO-style prompt every time, a team writes it once, names it, and lets the model invoke it when relevant. This pushes prompt engineering from per-message wording towards a library of small, well-scoped work specifications, each narrowing what the model attends to while keeping the discipline the role-as-lens view already requires: perspective, criteria, and responsibility.

\subsection{Instruction Design: Affirmative Targets and Guardrails}

The fourth design move is to phrase quality standards as targets instead of vague prohibitions. The principle is: give the model the target before listing mistakes to avoid. Affirmative instructions define the desired quality standard; negative instructions define hard boundaries, as Figure~\ref{fig:affirmative} shows.

\Figure[!t][width=0.48\textwidth]{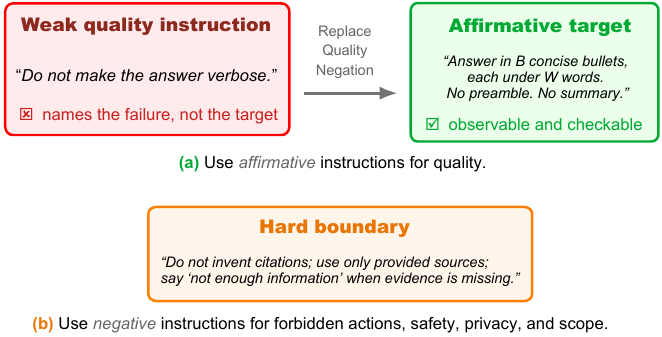}{Instruction design: (a) affirmative targets define observable behaviour; (b) negative instructions are best reserved for hard boundaries.\label{fig:affirmative}}

For example, ``Do not make the answer verbose'' names a failure but leaves the target underspecified. ``Answer in three concise bullets, each under twenty words. No preamble. No summary.'' creates an observable output requirement. Similarly, ``Do not hallucinate'' is less operational than ``Use only the provided sources; if evidence is missing, say `not enough information,' then ask the smallest useful set of questions that would clarify the grey areas.''

A related failure mode is the under-specified directive. A request like ``make it better'' names a direction but not an axis. Better in what sense: more accurate, more concise, faster to read, cheaper to run, easier to maintain, more persuasive, or more conservative? These axes are not independent, and they often pull against each other; gains in accuracy may cost tokens, gains in concision may cost coverage, and gains in safety may cost helpfulness. Without a chosen axis and an acceptance criterion, the request is effectively ill-posed: many valid improvements exist, several of them mutually exclusive, and reading between the lines does not help, because the missing information is not a hidden detail but a trade-off only the user can resolve. One general remedy is to elicit rather than only pre-specify: the assistant, or a dedicated skill, can be instructed to ask targeted clarifying questions until the axis, constraints, and acceptance criterion are pinned down, then act \cite{zhang2023clarify}. This turns specification from a solo drafting task into a short dialogue.

Negative instructions still matter for hard boundaries: do not invent citations, do not expose private data, do not modify files outside a permitted directory, and do not execute destructive commands (for example, commands that delete or overwrite data) without confirmation.

\paragraph{Design check.}
If an instruction says ``do not,'' ask whether it defines a hard boundary or only states a quality preference. Convert quality preferences into affirmative, observable targets.

\subsection{Occam's Razor and Chekhov's Gun}

The fifth design move is to make every prompt element earn its place. Occam's razor is a rule of simplicity: when two options work equally well, prefer the one with fewer parts and fewer assumptions \cite{wikipedia_occam,domingos1999occam}. In prompt engineering, it does not mean ``short prompts are better.'' It means unnecessary complexity is harmful. Chekhov's gun is a rule from storytelling: anything introduced into a story should matter later, and anything that does not should be cut \cite{wikipedia_chekhov,britannica_chekhov}. In prompt engineering, it means every instruction, constraint, role, or example left in the prompt should change the output or the workflow.

\Figure[!t][width=0.45\textwidth]{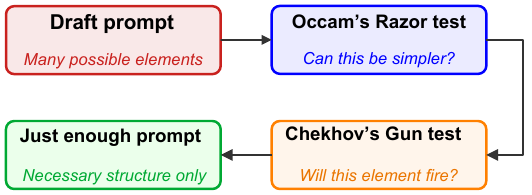}{Occam's razor and Chekhov's gun in prompt design: keep elements that shape behaviour, reduce ambiguity, prevent failure, or improve verification; cut decorative complexity that never fires in the output or workflow.\label{fig:occam-chekhov}}

The distinction, shown in Figure~\ref{fig:occam-chekhov}, is between necessary elements and decorative complexity. Necessary elements clarify the work: goal, input, audience, output format, constraints, examples, and checks. Decorative complexity adds role-play, vague emphasis, long preambles, or unnecessary multi-agent setups without improving the result. This principle is especially relevant in agentic workflows, where unnecessary tool calls, multi-agent debates, or long operating loops can add cost, latency, brittleness, and debugging difficulty. Edge cases need particular restraint: one exception often leads to another, and a prompt that tries to cover every unusual case loses the big picture, hiding the main instruction among rules that rarely apply. It is better to add an exception later, once the core prompt works and a real failure has shown that it is needed. A related hazard appears once guidance is spread across several sources: a global rule, a project-level rule, a skill, and the current request can point in different directions, and the model then has to settle the conflict, sometimes in a way the user did not intend. Each instruction should therefore be judged not only on whether it earns its place, but on whether it contradicts another.

\paragraph{Design check.}
After drafting a prompt, mark each element as goal, context, constraint, output requirement, example, critique step, tool rule, or verification step. If an element has no function, remove it.

\subsection{Critique, Steelmanning, and Premortems}

The sixth design move is to add critique when the cost of error justifies it. Because assistants are designed to be helpful, they may make weak plans feel stronger than they are \cite{sharma2024sycophancy}. The better response is structured criticism, not hostility.

Such criticism begins before the criticism itself. Steelmanning is the practice of addressing the strongest reasonable version of a position before evaluating it, the constructive counterpart of the straw man fallacy, in which a position is restated in a weaker form so that it is easy to attack \cite{aikin2011strawmen}. Before attacking an idea, the model should remove superficial noise and restate the argument in its strongest fair form. This stops surface defects, such as typos or weak phrasing, from substituting for real evaluation. After steelmanning, critique can focus on assumptions, evidence, trade-offs, risks, and failure modes. Table~\ref{tab:cfo} shows the corresponding instruction in the running example.

This practice has a limit worth stating in the prompt. Making a position stronger than the one its author actually holds is itself a recognised error, the \textit{iron man} \cite{aikin2016ironmen}. Asked to steelman a weak plan, a model may quietly fix its gaps and then judge the fixed version, approving a plan nobody proposed. Returning to the running example of Table~\ref{tab:cfo}, it might invent a hiring schedule the plan never gave, then judge staffing risk to be under control, hiding the gap the review should have reported. The prompt should therefore ask for the strongest version of the plan that its own text supports, and require any repair to be listed separately.

Critique still judges the plan on its own terms; a premortem \cite{klein2007premortem} changes the question. Instead of asking ``Is this a good plan?'', the prompt assumes that the plan has failed and asks what went wrong. A useful premortem asks the model to reconstruct the failure story and its early warning signs, then name the most likely failure, the most dangerous one, the biggest hidden assumption, and a revised direction. The purpose is to counter overconfidence by making failure the premise. The two moves are independent, but a premortem works better after steelmanning: a failure story told about a weakened plan finds faults the plan does not have.

Criticism can also be aimed at the model's own answer rather than at the plan, and repeated as a loop. In iterative self-feedback, the model drafts an answer, critiques its own output, and revises \cite{madaan2023selfrefine}; in verification-style prompting, it writes checking questions, answers them independently, and fixes inconsistencies before finalising \cite{dhuliawala2023cove}. Both reduce errors on tasks where a single pass is unreliable.

\paragraph{Design check.}
Use critique prompts when the task involves meaningful uncertainty, irreversible decisions, public communication, financial cost, safety, security, or complex implementation. For low-stakes rewriting, critique loops may be unnecessary or overkill.

\begin{table*}[!t]
\centering
\caption{The design moves applied to one task (a CFO plan review). Each move turns a weak prompt into a stronger specification; holding the task fixed isolates the contribution of each move. Symbols in the first column give the element of specification~\eqref{eq:spec} that the move instantiates.}
\label{tab:cfo}
\begin{tabular}{p{0.19\textwidth}p{0.36\textwidth}p{0.36\textwidth}}
\toprule
\textbf{Design move} & \textbf{Weak prompt} & \textbf{Strong prompt} \\
\midrule
Name the work ($G$) & ``What do you think of this plan?'' & ``Review this expansion plan and identify the assumptions that most affect the approval decision.'' \\
Build common ground ($C$, $I$) & No background supplied. & ``Context: a fixed \$5M, three-year SaaS expansion; the audience is the board.'' \\
Choose the lens ($L$, role) & ``Act as an expert CFO.'' & ``Review as a CFO accountable for cash discipline, downside risk, implementation cost, and measurable ROI.'' \\
Define the target ($K$, $F$) & ``Don't be vague.'' & ``List the five riskiest assumptions; for each, give its impact and one mitigation. Use only figures given in the plan.'' \\
Add critique ($L$, reasoning) & ``Is this a good plan?'' & ``Steelman the plan, then run a premortem: assume approval led to failure in 18 months and name the most likely cause.'' \\
Verify and escalate ($V$, $E$) & Accept the answer as written. & ``Check each figure against the plan, mark unverified numbers, and ask the user before recommending.'' \\
\bottomrule
\end{tabular}
\end{table*}

\subsection{Agentic Prompt Anatomy}

The final design move is to distinguish prompts that request an answer from prompts that authorise action. Agentic prompting extends the prompt from answer request to operating loop. An agentic prompt keeps every element of the prompt specification and adds only what execution requires: action boundaries, a tool set, a plan, and reporting. This extension can be written as

\begin{equation}
A = \langle P, B, T, \Pi, R \rangle,
\label{eq:agent}
\end{equation}
where $P$ is the prompt specification above, $B$ is action boundaries, including stop rules, $T$ is the available tool set, $\Pi$ is the plan the agent must set out, before acting or as it goes, so its intended steps can be checked, and $R$ is reporting: what the agent must report, and when. Goal, context, verification, and escalation are not repeated, since $P$ carries them.

\Figure[!t][width=0.48\textwidth]{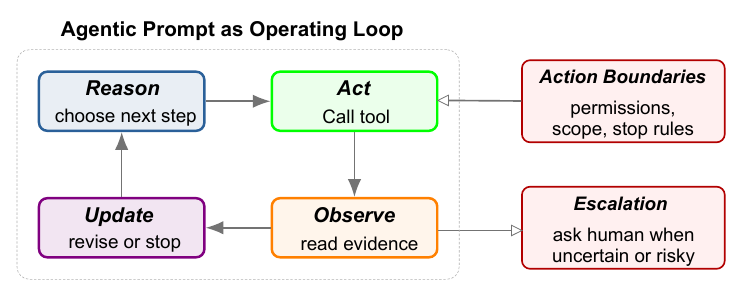}{Agentic prompting specifies an operating loop. Tool use is bounded by permissions, observations, stop rules, escalation, and reporting.\label{fig:agent-anatomy}}

For tool-using systems, a practical loop follows the ReAct pattern \cite{yao2023react}: reason, act, observe, update, as shown in Figure~\ref{fig:agent-anatomy}. The prompt should define what tools are for, what evidence the agent must collect, when it should stop, and when it should ask a human rather than guess. Tool sets are increasingly exposed through standard interfaces such as the Model Context Protocol (MCP) \cite{anthropic2024mcp}, which defines a common way for tools to describe what they do and to be invoked, while the prompt still decides when they should be used.

In such loops, the answer is often consumed by another program or another agent rather than read by a person. The output format $F$ must then be machine-readable: named fields with defined types and permitted values, instead of free-form text. In the CFO example, rather than a written list, the prompt can require one record per assumption with the fields \texttt{assumption}, \texttt{impact} (one of low, medium, high), \texttt{mitigation}, and \texttt{figure\_verified} (true or false). The receiving program can then check that five records were returned, as Table~\ref{tab:cfo} requested, that every \texttt{impact} value is permitted, and that no record leaves \texttt{figure\_verified} unset, so verification becomes automatic. The gain has a cost: constraining the output format can reduce the quality of the reasoning behind it, and tighter constraints tend to cost more \cite{tam2024formats}. The effect is uneven: answers the model must derive suffer most, while answers it merely selects from a fixed set of options are affected little. The practical rule is to let the model finish reasoning before it commits to an answer, either by giving the format a reasoning field placed before the answer it explains, or by taking two steps: the model answers in free-form text, and a second pass converts that answer into the required format.

\paragraph{Design check.}
If the model can act through tools, specify what it may do, what it must not do, how to interpret observations, when to stop, when to escalate to a human, and what it must set out: its plan before acting and its report afterwards.

\subsection{Worked Example: From Weak to Strong Prompt}

Table~\ref{tab:cfo} shows how the design moves work together. It takes one task, a CFO reviewing a business plan, and applies the moves one at a time, turning a weak prompt into a stronger specification. Keeping the task fixed shows what each move adds, which a single large example would hide.

Read top to bottom, the strong column builds into a single specification, while the weak column would stay one vague line. Figure~\ref{fig:cfo-prompt} shows the assembled prompt, with colour marking which design move each part came from.

\Figure[!t][width=0.48\textwidth]{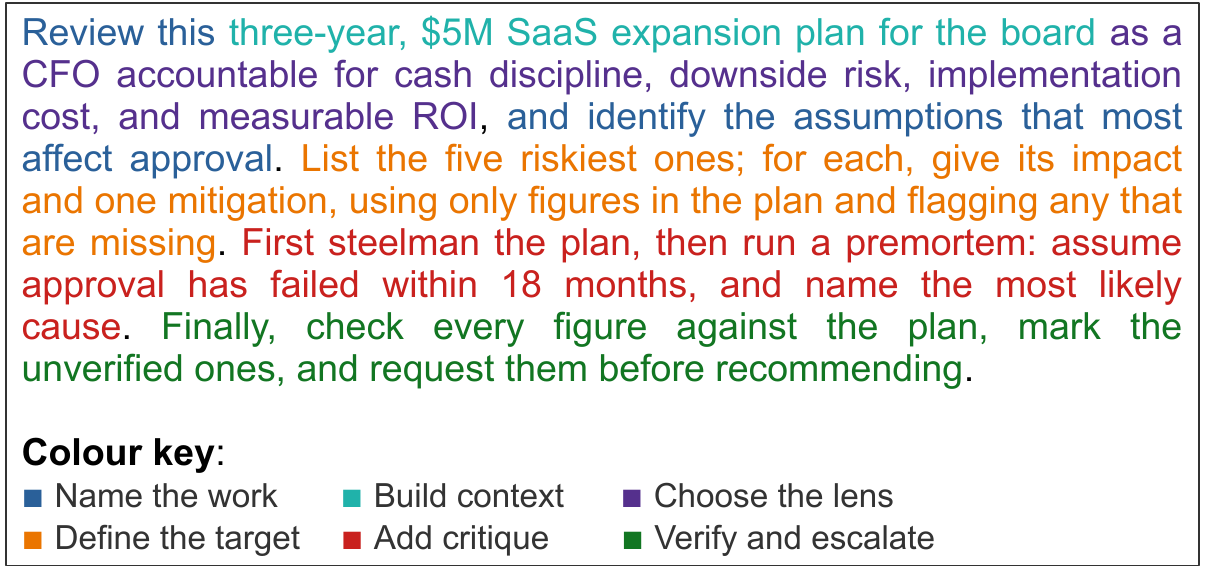}{The strong column of Table~\ref{tab:cfo} assembled into one CFO prompt; colour marks which design move each part came from.\label{fig:cfo-prompt}}

The six rows fill in every element of specification~\eqref{eq:spec}, with $L$ carrying both the role and the reasoning pattern. This is what the notation is for: it lets a draft prompt be checked for missing elements rather than for missing wording. The table shows the moves this task needs. The operating loop does not appear, because the review uses no tools, and the fifth move applies to every row rather than forming one of its own. If the review were handed to an agent that opens the plan and checks the figures itself, the remaining elements of~\eqref{eq:agent} would appear: permitted actions and files ($B$), the tools it may call ($T$), the plan it must set out ($\Pi$), and what it must report back ($R$). Not every task needs every move; each earns its place only when it removes an ambiguity the weak version leaves to the model.

\section{Design Principles and Checklists}
\label{sec:synthesis}

The framework can be summarised as design principles and checklists for task-specific adaptation. Table~\ref{tab:principles} summarises the core principles.

\begin{table*}[!t]
\centering
\caption{Prompt engineering principles for daily LLM work.}
\label{tab:principles}
\begin{tabular}{p{0.23\textwidth}p{0.35\textwidth}p{0.35\textwidth}}
\toprule
\textbf{Principle} & \textbf{Core idea} & \textbf{Practical implication} \\
\midrule
Prompt as work specification & The prompt defines task, context, criteria, constraints, output, and review. & Treat prompts as work briefs. \\
Explicit choices & Ambiguity should be converted into goals, constraints, evidence requirements, trade-offs, and success criteria. & Reduce hidden assumptions before the model commits to a path. \\
Task focus & Clear prompts keep the model on the subproblems that matter. & Avoid side problems and downstream errors in long chats or workflows. \\
Role as lens & Role playing works when it changes perspective, criteria, and responsibility. & Use roles such as CFO, security reviewer, or customer advocate only when each role has distinct criteria. \\
Affirmative target & Positive quality instructions define the desired behaviour. & Use negative instructions mainly for guardrails and forbidden actions. \\
Occam/Chekhov & Within a prompt, every element must do work. & Remove decorative instructions; retain only elements that affect output or prevent failure. \\
Just enough specification & How much prompt to write is set by what an error would cost, not by taste. & Escalate from a direct prompt to a work brief, then to critique and verification, then to bounded action; drop structure when a task turns out to be routine. \\
Steelmanning & Criticism should address the strongest fair version of a position, not a weakened one. & Restate an argument at its strongest before judging it. \\
Premortem & Assuming failure has already happened surfaces hidden assumptions and warning signs. & Ask why a plan failed before committing to it. \\
Agent anatomy & Agents require goals, boundaries, tools, a plan, verification, escalation, and reporting. & Design the operating loop, not only the answer. \\
\bottomrule
\end{tabular}
\end{table*}

These principles also yield a practical sequence. First, the work is named, then context and output requirements are specified. Second, critique and reasoning patterns are added when the task warrants them. Third, prompts are extended into agentic workflows. Fourth, prompts are evaluated with rubrics, prompt logs, and test cases. A rubric is a short checklist or scoring guide for judging whether an output meets the intended standard.

\paragraph{Compact design checklist.}
For reusable or consequential prompts, the framework reduces to the following checklist:
\begin{itemize}[leftmargin=*,label=$\square$]
    \item Is the work itself named clearly?
    \item Is relevant context present, and irrelevant context cut?
    \item Are hidden assumptions converted into explicit choices?
    \item Does the role name perspective, criteria, responsibility?
    \item Are standards phrased as affirmative, observable targets?
    \item Are hard boundaries stated explicitly?
    \item Does every prompt element affect the output, reduce ambiguity, prevent a failure, or support verification?
    \item Where the stakes justify it, is critique or verification asked for, and is it clear what would count as a failure?
    \item For agentic tasks, are tool permissions, the plan the agent must set out, observations, stop rules, escalation, and reporting defined?
\end{itemize}

\section{Discussion}
\label{sec:discussion}

\subsection{Why Principles Matter More Than Templates}

The framework developed in Section~\ref{sec:implementation} carries a recurring message: prompt engineering should develop principles and habits of thought rather than fixed wordings and templates. This matters because LLM behaviour changes across models, providers, settings, and tool environments. A fixed template may become stale, but the underlying design question remains: what does the system need to know, do, check, avoid, and return? Recent commentary puts this more sharply: findings tied to one model or release date do not accumulate into lasting knowledge, and practice should be judged by principles that carry across model generations rather than by wordings that will not \cite{pieraccini2026prompting}.

The framing of this paper follows from the fact that principles outlast wordings. Prompt engineering is treated here as a design discipline for translating human intent into specifications an AI system can act on, not as a science of how language models produce their outputs. The contribution is a coherent set of principles and checklists rather than an explanation of why any particular wording succeeds. Questions about model internals, capability limits, or emergent behaviour belong to a different literature and are left to it; see, for example, surveys of methods for interpreting the internal structure of neural networks \cite{rauker2023transparent}, studies of where compositional reasoning breaks down \cite{dziri2023faith}, and work on how generalisation emerges during training \cite{bertolotti2026grokking}.

Many useful principles come from general problem-solving practice: Occam's razor, Chekhov's gun, steelmanning, premortems, role-based review, and explicit criteria. These principles are useful in prompt engineering because prompts are written in human language and are used to coordinate problem-solving behaviour. LLMs are not human, and the paper does not claim that they reason as humans do. The narrower claim is that these principles are encoded in language, examples, and professional discourse, so they work as communication structures when interacting with LLMs.

\subsection{Just Enough Specification}

Two boundaries limit what the framework claims. First, prompt structure is not model capability: better models reduce the burden of saying everything explicitly, but they do not remove the need to clarify the task, constraints, and review criteria. This is consistent with AI engineering practice, where prompts, retrieval settings, tool choices, and model changes are tested against explicit task criteria rather than judged by first impression \cite{huyen2025aiengineering,openai2026evals}. Second, the escalation ladder in Table~\ref{tab:escalation} exists so that the framework does not drift into advocacy for long prompts. The practical target is ``just enough specification'': enough structure to make the work reviewable and reliable, but not so much that the prompt becomes noisy, costly, or brittle.

\subsection{Boundaries}

Prompt engineering does not replace domain expertise, source verification, testing, privacy review, or security controls. It also does not make model behaviour identical across providers, versions, temperature (the randomness setting), tools, or deployment environments. The same prompt may behave differently when context windows, retrieval settings, tool permissions, or system instructions change; even meaning-preserving edits such as formatting or example ordering can shift results \cite{sclar2023sensitivity}. For that reason, reusable prompts and agentic workflows should be reviewed as living design artifacts rather than fixed scripts.

The same caution applies to the examples here. Each was chosen to show one distinction: persona against lens in the CFO review, a target against a prohibition in the affirmative-instruction example, and the way failure-first framing exposes assumptions in the premortem. They show what each move does, not how much it helps.

\subsection{Risks and Misuse}

Prompt engineering can create a false sense of control. A well-structured prompt may produce a polished answer that is still wrong. A role label does not verify the answer \cite{zheng2024personas}, longer reasoning is not necessarily better \cite{sprague2025cot}, and LLMs are not reliable sources of up-to-date information. In consequential settings, source checking, test cases, rubrics, and human review remain necessary.

Agentic systems add security risks. If an agent can browse, execute code, edit files, or call external services, prompt injection and untrusted content become workflow risks \cite{greshake2023injection}. The control principle is to separate trusted instructions from untrusted data, and to define action boundaries, least privilege, confirmations, and stop rules.

Those controls assume the threat arrives in material the agent fetches: a web page, a file, or a result returned by a tool. The prompt itself can also be the attack. A user writes wording that pushes the model past its own guidelines, for instance by assigning it a role that is exempt from them, or by instructing it never to refuse; this is known as \textit{jailbreaking} \cite{wei2023jailbroken}. Prompt design offers little defence here, since resistance rests on the provider's safety training and on filtering in the surrounding application. Prompts can also leak information: the model can be asked to repeat the instructions it was given, and, depending on the service, what is sent may be retained or used for training. Credentials, internal thresholds, and business rules therefore do not belong in a prompt \cite{owasp2025llm}. What reaches the model is also more than what the user typed. Applications add their own instructions, tool descriptions, and account or organisation details before sending a request, so a prompt can carry information its author never wrote and would not have chosen to disclose, which matters when the work itself is confidential.

\subsection{Implications for Everyday LLM Work}

In professional use, these design moves draw on skills people already use when handing work to a colleague: delegation, decision quality, stakeholder analysis, implementation risk, requirements specification, testing, tool use, and safety limits. Research on human-AI interaction points the same way, stressing user control, recovery when the system fails, expectations matched to what the system can actually do, and reliance proportionate to how far it can be trusted \cite{amershi2019guidelines}. Practice and research agree on how the work should be split: the user owns the problem, the approach taken to solve it, and the standard the answer is judged against, while the model works within that frame rather than setting it. Its output is a draft to be checked, not a verdict to be accepted, and its agreement is not evidence that the framing is right \cite{sharma2024sycophancy}. If the user does not state the task precisely, the model guesses what was meant and answers that guess. Nothing in the reply marks it as a guess, so the user can accept a polished answer to a task they never set. Done carefully, prompt engineering sharpens human thinking: the user must understand the problem, the standard, the constraints, and what failure would look like.

\section{Conclusion}
\label{sec:conclusion}

Prompt engineering is more than clever phrasing. As models move from answering questions to carrying out tasks, it is better understood as the design of work specifications for AI systems: the discipline of translating human intent into terms those systems can act on. On that view, a strong prompt makes intent, constraints, success criteria, and verification explicit.
Turning that view into practice is the paper's contribution, a learning path that moves from messy thoughts to AI workflows by turning ambiguity into explicit choices, treating roles, and moderated panels of them, as attention lenses, preferring affirmative quality targets while reserving prohibitions for hard boundaries, applying Occam's razor and Chekhov's gun so every element earns its place, adding structured criticism through steelmanning and premortems, and specifying agentic operating loops with explicit boundaries and escalation. The strongest prompt is rarely the longest prompt; it is the one that makes desired behaviour, required sources and checks, and success criteria unmistakable.

\section*{Generative AI Tools Disclosure}
The authors used generative AI tools for proofreading. All AI-assisted edits were reviewed by the authors, who are responsible for the content.

\bibliographystyle{IEEEtran}
\bibliography{ref}

\end{document}